# Solving Optimization Problems through Fully Convolutional Networks: an Application to the Travelling Salesman Problem

Zhengxuan Ling, Xinyu Tao, Yu Zhang, Xi Chen*

**Abstract**—In the new wave of artificial intelligence, deep learning is impacting various industries. As a closely related area, optimization algorithms greatly contribute to the development of deep learning. But the reverse applications are still insufficient. Is there any efficient way to solve certain optimization problem through deep learning? The key is to convert the optimization to a representation suitable for deep learning. In this paper, a traveling salesman problem (TSP) is studied. Considering that deep learning is good at image processing, an image representation method is proposed to transfer a TSP to an image. Based on samples of a 10 city TSP, a fully convolutional network (FCN) is used to learn the mapping from a feasible region to an optimal solution. The training process is analyzed and interpreted through stages. A visualization method is presented to show how a FCN can understand the training task of a TSP. Once the training is completed, no significant effort is required to solve a new TSP and the prediction is obtained on the scale of milliseconds. The results show good performance in finding the global optimal solution. Moreover, the developed FCN model has been demonstrated on TSP's with different city numbers, proving excellent generalization performance.

**Index Terms**— Travelling salesman, Deep learning, Optimization, Mapping, Fully convolutional network

## 1 INTRODUCTION

In recent years, we have witnessed a new upsurge in artificial intelligence represented by deep learning, which is beginning to truly integrate into many aspects of daily life. Practitioners from all walks are trying to combine deep learning

---

*Corresponding author, email: xi_chen@zju.edu.cn.

with different industries and have made great achievements in the fields of autonomous driving [1], [2] medical imaging [3], [4], machine translation [5], drug design [6] and so on. Deep learning has demonstrated similar or even better comprehension capability and problem solving ability than trained professionals in many industries.

Practitioners tend to have greater understanding and problem solving ability than the average person in their professional field because of their comprehensive professional knowledge. Only a small part of this knowledge can be taught to computers through regular programming languages and the rest may be difficult to describe in succinct terms. One solution is to let the computer program learn by itself. Through deep learning, algorithms can establish a complex mapping between historical data and knowledge concepts through many training data. For instance, when learning to recognize cars from images, deep convolutional network can extract many elementary features from the arrangement of pixels in different images, then features are combined to form a more general feature. Eventually the concept of car and the mapping between the car and some fixed pixel arrangement mode is established. The capability of forming complex mappings is the key to "intelligence" [7]. Once such a mapping is formed, people will be less bothered by the technical details of the intermediate link and speed up the solving process.

As a research field, optimization is closely related to deep learning. Optimization algorithms are crucial parts of deep learning. It is difficult to build deep neural networks without efficient optimization algorithms. We can even say that optimization algorithms are the driving force of deep learning's "learning" ability. A large body of research has focused on developing optimization algorithms for deep learning. However, the reverse study of applying deep learning to optimization has not been sufficient so far.

Mapping relations also widely exist in optimization. For a long time, people have worked to obtain non-iterative methods for optimization problems. To our knowledge, however, the attempts to use neural networks to find mappings between the parameters of optimization problems and their corresponding optimal solutions all failed to achieve ideal results. This is mainly because the mappings between the parameters and the optimal solution are too complicated for neural networks to fit, and insufficient training samples often lead to serious overfitting problems. Therefore, optimization algorithms use an iterative search strategy to solve optimization problems. Large numbers of iterations are needed to obtain the optimal solution, and once the parameters of problem are changed, the iterations need to be restarted.

However, in recent years, deep convolutional neural network (DCNN) structures such as AlexNet [8], VGGNet [9], and Resnet have been proposed, which greatly improve the fitting capability of neural networks. But there still exists an obstacle in applying these structures to optimization problems: the forms of input and output data that can be accepted by DCNN's are still limited, such as matrices and vectors. If the parameters of the optimization problem are forcibly converted to these forms, it is difficult to extract effective features through convolution operations.

It is well known that DCNN's are good at processing images. Thus, is it possible to characterize optimization problems as images and convert the parameters of the optimization problems into a form that can be easily extracted by convolution operations? In this paper, a positive answer is given by taking traveling salesman problem (TSP) as an application. An image representation is proposed for TSP. The parameters of the problem are characterized in the form of points and lines which can be easily processed by neural networks. Once the training process is completed, the neural network can generate solutions for a new TSP within milliseconds. Meanwhile, high solution accuracy is observed.

## 2   RELATED WORK on Travelling Salesman Problem

The TSP is a typical NP-hard integer programming optimization problem which has a wide range of applications in science and industry [11]. The problem can be described as a salesman who needs to visit a set of different cities and return to the departure city. One needs to determine the order of the cities such that each city is visited exactly once, and the goal is to minimize the total distance travelled [12]. The optimization formulation of a TSP is shown in equation (1). The set of cities is denoted as $V$; $Z$ is the total distance; $d_{ij}$ is the distance of the path between the $i^{th}$ city and $j^{th}$ city; $t_{ij}$ indicates whether the path belongs to the loop or not, and $S$ is a subset of $V$. The first two constraints imply that each city can be passed only once, and the third constraint guarantees that no sub loop solutions exist.

$$\min Z = \sum_{i=1}^{n}\sum_{j=1}^{n} d_{ij} t_{ij}$$

$$s.t. \begin{cases} \sum_{j=1}^{n} t_{ij} = 1, i \in V \\ \sum_{i=1}^{n} t_{ij} = 1, j \in V \\ \sum_{i \in S}\sum_{j \in S} t_{ij} \leq |S| - 1, \forall S \subset V, 2 \leq |S| \leq n-1 \\ t_{ij} \in \{0,1\} \end{cases} \quad (1)$$

At present, most algorithms for the TSP tend to search and iterate among all paths that meet the constraint conditions. The algorithms can be classified into deterministic algorithms and heuristic algorithms [13]. Common deterministic algorithms include exhaustive search, branch and bound, dynamic programming, etc. However, the number of feasible solutions and the number of customers grows exponentially. Heuristic algorithms, such as genetic algorithms, ant colony algorithms or simulated annealing etc., normally take many iterations to obtain an acceptable approximate solution.

Neural networks have also been applied to the TSP. The appropriate energy function was introduced in Hopfied neural network [14] to make it consistent with the objective function of a TSP. The energy continuously reduces with iteration, and finally reaches equilibrium, indicating a local optimal solution. Irwan Bello et al. used reinforcement learning combined with deep neural networks to solve the TSP [15]. Li et al. trained the Graph Convolutional Networks to improve the search speed of Guided Tree Search [16]. These methods adopt search strategy which cannot directly establish the mapping to the final solution. Thus, a new iterative process is needed when dealing with a new problem. In contrast to these methods, we establish the mapping *F* between the cities' coordinate set *(X, Y)* and the optimal city passing order, denoted as *{C1, C2, ..., CN}*, as shown in equation (2).

$$\{C_1, C_2, \ldots, C_N\} = F(X, Y) \quad (2)$$

To find out the mapping, a fully convolutional network [17] is adopted as shown in Fig. 1. The coordinates of the cities, which are generated randomly, will be converted to images to be input to the FCN. The training labels are the solutions of the TSP as produced by dynamic programming.

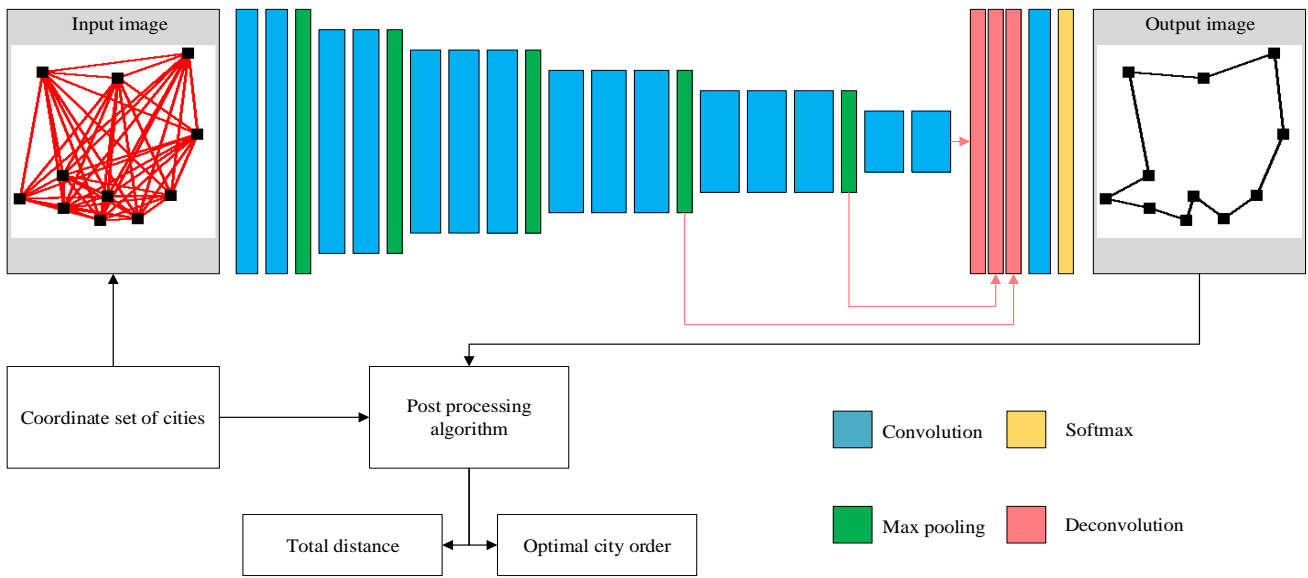

**Fig. 1. System architecture**

## 3 SYSTEM ARCHITECTURE

### 3.1 Representation

As mentioned before, the mapping established by the neural network is based on many training data. Thus, it is quite important to generate a batch of TSP samples with different parameters and represent these samples in a proper way. To this end, we represent the coordinate set of cities as a fully connected graph. If fully connected graph is adopted to be the input of FCN, the rectangular convolution kernel should be replaced with a kernel that is suitable for the graph. This may limit the range of using neural network since the kernels need to change to adapt different graph structures [18]. Therefore, the graph is projected to a fixed size image. Image is kind of a Euclidean structure. The advantage of using Euclidean structure is that the pixels are arranged neatly so that the convolution operations can easily be applied to extract features.

Before generating the fully connected graph, normalization is needed to adjust the coordinates of cities into the same scale. This is because the scale of distance among different TSP varies largely, it is quite difficult to accommodate all these graphs using images with fixed size. Through normalization, graphs can be projected to the center of an image and fill the whole image. This can eliminate the noise caused by displacement and reduce the imbalance of pixel numbers between background and image. The relations between the coordinates of cities and their positions on the images are shown in equation (3).

$$\begin{cases} \lambda_x = \frac{x_{max}-x_{min}}{w}, \lambda_y = \frac{y_{max}-y_{min}}{h} \\ x_i^* = \frac{x_i-x_{min}}{\lambda_x}, y_i^* = \frac{y_i-y_{min}}{\lambda_y} \\ i = 1,2,\dots,n \end{cases} \tag{3}$$

In equation (3), $x_{max}$ and $y_{max}$ respectively represent the maximum horizontal and vertical coordinates among all the cities while $x_{min}$ and $y_{min}$ respectively represent the minimum horizontal and vertical coordinates. w and h are the length and width of the image. $\lambda_x$ and $\lambda_y$ are scaling coefficients. $x_i$, $y_i$ are coordinates of the $i^{th}$ city and $x_i^*$, $y_i^*$ are the coordinates of the $i^{th}$ city on the image. The integer part of $x_i^*$, $y_i^*$ is the position of the $i^{th}$ city.

It should be noted that the image, as a kind of bitmap, should be in high resolution. If the resolution is lower than a certain threshold, image information will be lost in the conversion process, which will affect the judgment of the neural network. For example, the paths among cities will have serious jaggy distortion. When two cities are very close, most of the paths overlap and are unrecognizable. An image with high resolution can not only make the paths more uniform and clearer, but also can accommodate a TSP with more cities. Therefore, a 224×224 dimension, 3-channel RGB image is employed as input image.

After projection, each city only occupies one pixel on the image, which makes it difficult to figure out the position of cities. Moreover, after joining the paths, the information of cities is easily covered by the paths, which may cause FCN identify the positions of cities incorrectly. In view of this situation, the following adjustment of the image is conducted:

1. Pixels that are within 6 pixels horizontally and vertically of each city location are changed to the same color as the city.

2. Paths adopt a different color from cities.

For building the neural network, a FCN is adopted rather than a general DCNN. The goal of the neural network is to find the optimal path among all possible paths in images, but general DCNN's are good at giving image-wise judgment, such as if an image contains a cat. This kind of judgment provides too little information to find the solution of a TSP. The FCN approach can avoid this defect due to its pixel-wise classification capability: each path in image consists of pixels, and each pixel will be given a judgment on whether it belongs to the optimal path or belongs to the background. Paths are judged according to the results on pixels through a voting mechanism.

The dimensions of a FCN's output are consistent with its input. Considering this characteristic of FCN's, the label image should have the following characteristics:

1. The length and width of the label should be consistent with the input image.

2. The optimal path should be contained in the label image.

3. The label has a one-to-one mapping with each pixel of the input image so that the FCN can classify every pixel according to the label.

A set of TSP's are solved by a deterministic algorithm to obtain accurate labels. Then we draw the cities on a two-dimensional plane and connect the cities according to the optimal cities' passing order to form the optimal path graph. Similarly, equation (3) is used to transfer the graph into an image with the same length and width as the input image. Subsequently, we binarize the image by setting the value at each pixel as 0 or 1, where 1 indicates that we want FCN to classify the pixel as the position of the optimal path and the cities, and 0 indicates that FCN is expected to put this pixel into another class. To facilitate pixel classification, One-hot encoding [19] is also required to transfer the label into a matrix of 224×224×2.

### 3.2 Fully convolutional network

FCN is adopted based on the VGGNet [9]. If a neural network with fewer layers is selected, insufficient fitting and overfitting problems may occur. A FCN retains the first five convolutions and pooling operations of the VGG structure. The size of the input image will be reduced by 2, 4, 8, and16 times in the five convolution and pooling operations; it eventually becomes a feature map with the size of 7×7×1024. In contrast to a VGG structure, the FCN structure uses a deconvolution layer to make the output image of the network the same size as the input image, thus the feature map is required to be up sampled by 32 times. Since the pixels in the 7×7×1024 feature map each have a large receptive field, deconvolution of the feature map alone will result in insufficient detail. In contrast, the feature maps of the previous pooling layers have more detailed information; but generalized features are rarely found in the feature maps. Therefore, these feature maps can complement each other. To this end, the feature maps output by the fourth and third pooling layers are deconvolved simultaneously with 16 times and 8 times up-sampling, respectively.

After up-sampling, the three feature maps are the same size. We concatenate them together and form a matrix of 224×224×3. For classification, the matrix needs to be converted to 224×224×2, which is accomplished by a convolution operation. Compared to adding all the feature maps after the up-sampling, concatenation and convolution operations can provide more abundant ways of combing the features [4]. After convolution and concatenation, a sigmoid function is used to map all values to probability values. After the conversion, there are two probability values at each pixel position in consistency with the label. One represents the probability that the pixel belongs to the optimal path, and another represents the probability of belonging to the background.

The loss function of a FCN is used to measure the difference between the probability values on the output of FCN and labels. The goal of the optimization is to minimize the loss function. Equation (4) shows the loss function of FCN where y is the probability distribution of the network's output and y´ is the probability distribution of the label.

$$J_{y'}(y) = -\sum_i^w \sum_j^h \sum_k^2 y'_{ijk} \log(y_{ijk})/(2wh) \qquad (4)$$

To better illustrate the classification results of the FCN, it is necessary to convert the 224×224×2 matrix into an image. We compare the two probability values on each pixel and classify the pixel into the class with the larger probability value.

Then we color the picture according to the result of the classification. Pixels that belong to the optimal path class are in black and the rest are in white.

### 3.3 Post processing

As shown in Fig. 1, a post processing block is required to retrieve information obtained from the FCN output. In the output picture from the FCN, the density difference between different paths is distinguishable. Starting from any city and moving forward according to the path with the highest density of the black pixels, one can always return to the departure city that meets all the constraints of TSP. However, the solution cannot be extracted by computer yet. In addition, due to misclassification of pixels, there are little portions of the FCN's output image that have broken or redundant paths. When this problem occurs, one can hardly identify the solution according to the output image alone.

---

**Algorithm 1** Post Processing Algorithm

---

1: Input: Coordinates sets of cities; Departure city;
2: Randomly select $m$ cities from $n$ cities
3: Take the m cities as departure cities in turns, denoted as $C_1$
4:     Mark $C_1$ as the current city
5:     for $i=1, ..., n-1$
6:         Make paths between the current city and the remaining $n-i$ cities and then compute $\xi$
7:         Choose the city that has biggest $\xi$ to be the current city and mark it as $C_{i+1}$
8:         Compute the distance between $C_i$ and $C_{i+1}$
9:     Compute: the distance between $C_n$ and $C_1$
10:     Compute: the optimal city order; the overall distance
11: Compare the overall distance among $m$ departure cities; choose the smallest one
12: Adjust the optimal city order according to the given departure city
13: Output: Overall distance; Optimal city order

---

To automatically detect the solutions produced by FCN and generate acceptable solutions when problems mentioned above occur, a post processing algorithm is developed. The algorithm can calculate the density value of black pixels on different paths and output uniquely determined solutions in the form of total distance and the cities' passing order. The proposed post processing algorithm is briefly summarized as algorithm 1.

The main strategy of the algorithm is to search the paths with the highest black pixel density city by city and finally form a loop. To calculate the density value, the algorithm samples the pixels on the path. Denote the path between the $i^{th}$ city and the $j^{th}$ city as $L_{ij}$. The number of sampled pixels on the line $L_{ij}$, denoted as $p_{ij}$, is the larger of the horizontal and vertical coordinates' absolute difference between the two cities.

$$p_{ij} = max < |x_i^* - x_j^*|, |y_i^* - y_j^*| > \tag{5}$$

$$\begin{cases} x_t' = [min < x_i^*, x_j^* > + |\frac{t \cdot (x_i^* - x_j^*)}{p}|] \\ y_t' = [min < y_i^*, y_j^* > + |\frac{t \cdot (y_i^* - y_j^*)}{p}|] \\ t = 1, 2, \ldots, p \end{cases} \tag{6}$$

$$\rho_{ij} = q_{ij}/p_{ij} \tag{7}$$

where $x^*i, x^*j, y^*i, y^*j$ are the horizontal and vertical coordinate values of the $i^{th}$ city and the $j^{th}$ city, respectively. By taking the maximum value of the two, the algorithm can sample as many pixels as possible. Through this method, the calculated density value can be as close as possible to the true value. The coordinates of the points of the sampled pixels on the $L_{ij}$ are shown in equation (6). $x'_t$ and $y'_t$ are the horizontal and vertical coordinate values of the $t^{th}$ sample pixels. *[x]* obtains the integer part of x. If $q_{ij}$ pixels of these $p_{ij}$ pixels are black, then the density of black pixels can be calculated as shown in equation (7).

Due to misclassification by the FCN in some cities, the order of cities processed by the algorithm will affect the accuracy. For instance, when the misjudgment occurs in city $C_i$, the algorithm may make a mistake in choosing the next city. When dealing with the remaining *n-i* cities, the possibility of making incorrect choice still exists since the cities that have been passed before are no longer taken into consideration. Therefore, a deviation between the result of the algorithm and the FCN's output continues to accumulate. If $C_i$ could be judged in the end, the deviation could be minimized. The departure city is often given by the problem statement of the TSP, which means that the order of cities processed by post processing algorithm is fixed. But which city to start from does not matter in deriving the optimal path loop. Thus, m cities are selected to be the departure city in turns in the post processing algorithm to obtain a suitable processing order.

## 4 TRAINING PROCESS

Since the TSP does not have a unified large scale sample set for training and testing, we first generate a sample data set. The 10 city TSP is first studied in this project. The coordinates of 10 cities are generated randomly and converted into fully connected images through the method described in section 2.1. To make sure optimal solutions are used to generate the labels, the deterministic dynamic programming is used to solve the TSPs. The hardware platform is NVIDIA RTX 2080ti GPU and Intel 9700K CPU.

### 4.1 Parameter settings

Before training, the weight and bias parameters w and b of the neural network are initialized and training parameters are given. *w* is initialized by Xavier, which has been proved [20] to make the variance of the output of each layer in the neural network to be approximately equal, thus making the information propagate through the network more smoothly. Bias *b* is assigned to be 0.1. To improve the capability of expressing nonlinearity, the network takes Relu as an activation function. Adaptive moment estimation (Adam) is adopted to be the optimization function with a learning rate $\alpha=10^{-4}$.

### 4.2 Overfitting

Overfitting is a key issue during the training process. A variety of methods are used to prevent this problem.

**Dropout** [21] could reduce the complex co-adaptive relationship among neurons. The updating of weights no longer depends on the interaction of nodes with fixed relations. We set the ratio of dropout to 0.5 during the training.

**Truncated Training** can prevent further development of overfitting. 3480 TSP samples are generated, among which 3000 are used as training data and 480 as test data. As shown in Fig. 2a, after 3,000 iterations, the difference increases rapidly, indicating a worsening overfitting phenomenon. Thus, we truncate the training at 3,000 iterations.

**Increasing Training Samples** is also an important means to reduce overfitting. We expanded the training dataset to 21,000 samples and take 3,000 samples as a batch. After every 3,000 iterations, a new batch of training data is added. The effect of adding training samples is evaluated by observing the learning curve. As can be seen from Fig. 2b, the loss function curve of the training dataset and the test dataset decrease at the beginning of each new sample set introduction. When the size of training dataset reaches 9000, the effect of adding new samples is negligible. Therefore, 9,000 samples are eventually used for the training.

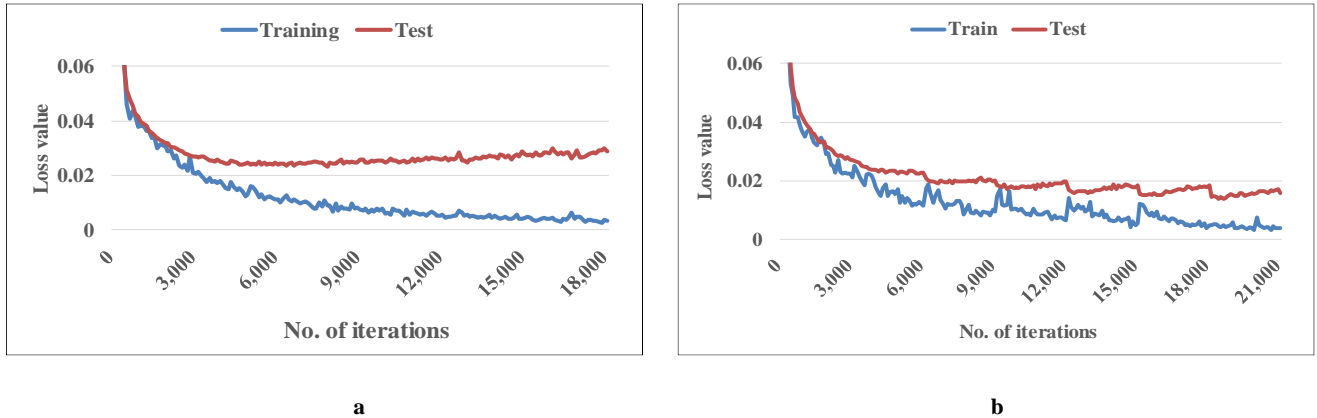

**Fig. 2. Learning Curve of the Loss Function. a** is a learning curve using 3,000 training samples while **b** uses 21,000 training samples with dynamic training. The blue line is the loss function curve of training dataset, the orange line is the loss function of test dataset.

### 4.3 Training process visualization

The human brain constantly changes the connection between neurons during a learning process. Similarly, the artificial neural network can combine the information extracted from the input data and constantly change the patterns of information composition during the training process. The change of patterns is reflected by the change in the weight parameters w and the bias parameters b. Unfortunately, neural networks are not interpretable yet, people still cannot understand how artificial neural networks learn to solve problems. But in the field of optimization, optimization algorithms are generally interpretable. For most algorithms for solving the TSP, the search strategy is clear. For example, all possible path schemes are listed by exhaustive method for comparison, and genetic algorithms adopt mutation and selection strategies. In the process of comparison or selection, the TSP problem will find a more optimized result step by step.

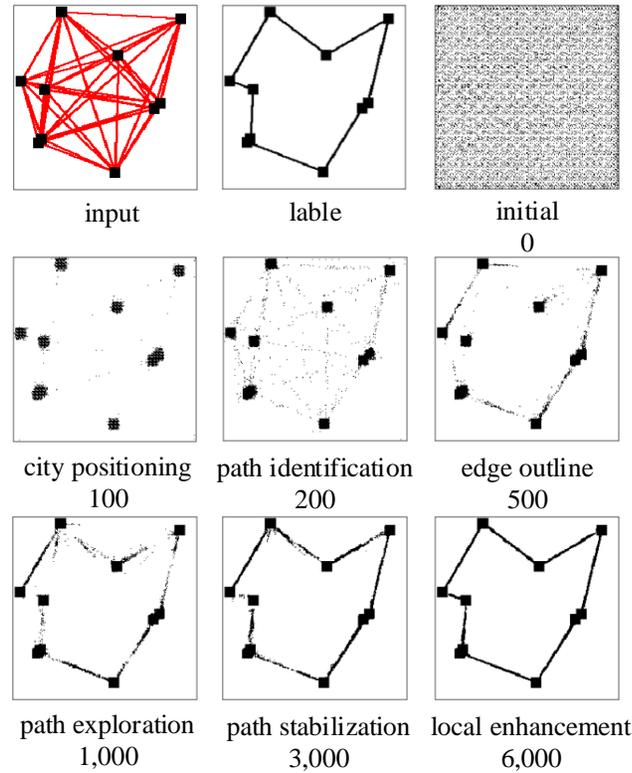

Fig. 3. The output of FCN at various stages of training process.

A visualization method is adopted to observe the training process of the FCN. We find that the FCN shows considerable interpretability during training, in which the network continuously adjusts its mapping mode to approach the true mapping relationship between input and output.

The visualization method we used is to input a batch of samples to the FCN after every 50 iterations during the training process. We then record the output of the FCN. Based on observation and comparison of these samples, the FCN training process can be roughly divided into: an initial stage, a city positioning stage, a path identification stage, an edge outline stage, a path exploration stage, a path stabilization stage and a local enhancement stage. As shown in Fig. 3, besides the input and label images, a series of snap results at different iteration stages are presented, where the number below each subplot denotes the iteration number of the snap results. In Fig. 3, a typical test sample is chosen as an example to show the training process of FCN. Let's look at what the FCN does at each training stage.

**Initial stage**: At this stage, the iteration has not yet started; the black pixels in the output picture are evenly distributed over the entire image, validating that with the initial parameters, the network cannot retrieve useful information from the input image.

**City positioning stage**: After 100 iterations, there are higher density of the black pixels at the positions of the cities than other positions in the image. This phenomenon indicates that the neural network first identifies the city information from the input image. This is mainly because the location information of the city in the input image is one-to-one mapping to the label. However, during this stage the locations of the cities are not exactly square as the cities shown in the input images. FCN will continue adjusting the shape of the cities' locations in the next few stages.

**Path identification stage**: After 200 iterations, the locations of paths is roughly outlined, indicating that the neural network has been able to retrieve the path information from the input image. In the positions where the paths exist, some pixels are black, but the network cannot distinguish the optimal path yet; thus, all possible paths are given similar probability values.

**Edge outline stage**: Some cities are located around the edge of the image, while others are relatively closer to the center. After 500 iterations, the density of black pixels on the paths among edge cities rises rapidly. In contrast, the density of the paths among center cities rapidly decreases. The neural network seems to generate an initial solution of a TSP at this stage by connecting the edge cities to form an outline. This "outline" solution is not a feasible solution, as it does not pass all cities. But the solution is not strange. It is very similar to the behavior of human beings, as many people will also first identify the outline of the cities when attempting to solve the TSP manually.

**Path exploration stage**: At previous stage, the neural network increases the likelihood that the outermost connection would be the optimal path. However, this is not the final solution given by the FCN. After 1,000 iterations, the neural network adopts a heuristic strategy, which repeatedly increases the density of black pixels of the paths that connect the edge cities and center cities, and simultaneously reduces the density of other adjacent paths. By this strategy, the two cities located in the middle part of the image are also considered by the neural network. Feasible solutions are generated continuously, and the network will also constantly compare these feasible solutions.

**Path stabilization stage**: After 3,000 iterations, the density of black pixels is no longer randomly changed. Rather, the change of density has directionality. The density values of some paths start to rise rapidly while the density values of other paths decrease rapidly, which state that the FCN has determined the final result from these feasible solutions.

**Local enhancement stage**: This stage starts after about 6,000 iterations. At this stage, the neural network no longer adjusts the output significantly. Instead, the FCN fine tunes the shapes of cities and paths to make them more regular and more similar to the data labels. This stage will last until the end of training.

Similar to the sample shown in Fig. 3, the training process for other samples can also be categorized into these stages. The whole training process of the FCN is a little bit like a traditional optimization process. Until the edge outline stage, the FCN selects an initial solution; during the path Exploration stage, the FCN searches for feasible solutions and compares them; the neural network selects the final result during the path stabilization stage and continues to adjust local details during the local enhancement stage.

The dynamic visualization results can help to understand how the FCN learns. It should be noted that the sample shown in Fig. 3 is never learned by the FCN and does not need to be learned again. The visualization is generated on a test sample by storing the parameters from during training and applying these partially trained parameters on the new sample. Validation with a new sample reveals that a general mapping to the optimal solution has been learned by the FCN.

## 5 RESULTS AND DISCUSSION

After training, we input 480 sets of test samples into the network and analyze the output images for validation. To comprehensively quantify the solution qualities of the FCN network, we establish the following metrics:

1. $e_0$: Proportion of samples that the output distance optimized by the proposed method **equals to** the label distance.

2. $e_1$: Proportion of samples that the output distance optimized by the proposed method doesn't exceed **1.01** times the label distance.

3. $e_2$: Proportion of samples that the output distance optimized by the proposed method doesn't exceed **1.02** times the label distance.

4. $e_5$: Proportion of samples that the output distance optimized by the proposed method doesn't exceed **1.05** times the label distance.

5. $e_{10}$: Proportion of samples that the output distance optimized by the proposed method doesn't exceed **1.1** times the label distance.

6. $R_{aver}$: Average ratio of the overall distance output by the proposed method to their optimal path's overall distance

For 480 test samples, the calculated metrics are listed in Table 1. The optimal result is obtained in 86.88% of the samples. In addition, 95.42% of the samples can obtain a result that the total distance of the solution is less than 1.01 times the distance of the optimal solution. If the deviation tolerance is set to 5%, then all the solutions meet the requirement. The overall average ratio is 100.16%, that is, our proposed method only produces 160m of deviation for every 100km on average. Moreover, as there is no need for iteration, our method is suitable for a variety of real-time solving requirements.

**Table 1 Metrics on test samples**

| $e_0$ | $e_1$ | $e_2$ | $e_5$ | $e_{10}$ | $R_{aver}$ |
|---|---|---|---|---|---|
| 86.88% | 95.42% | 98.12% | 100.00% | 100.00% | 100.16% |

Under most circumstances, the FCN can clearly give accurate results. But some extreme samples will cause challenges for the FCN. These samples are particular because of two characteristics: the path loop is in an obviously concave shape; several cities are very close to each other. To illustrate the performance of the FCN, four samples are presented in Fig. 4.

Among these samples, **a** and **b** are ordinary samples while **c** and **d** are extreme samples. To facilitate comparison, we converted the cities' passing order output by the post processing algorithm into an image: The blue dots represent the locations of the cities; the path colors denote the traveling order, which are red, orange, yellow, green, cyan, blue, purple, black and olive in a sequence.

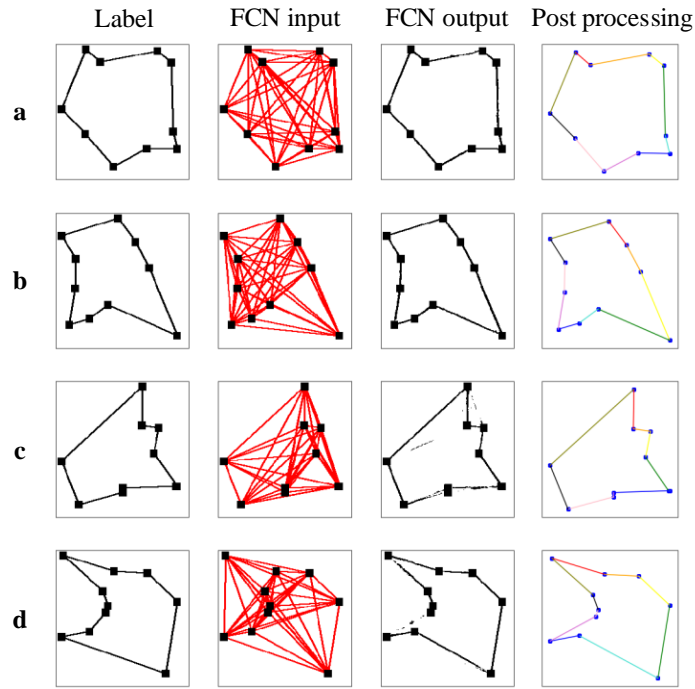

**Fig. 4. Results of the proposed method on new samples. a** and **b** are ordinary samples while **c** and **d** are extreme samples.

In the first two samples, most cities are edge cities, so it is not hard for the FCN to derive the path loop. The path loop derived by the FCN is clear. In subplot c, however, edge cities and center cities occur alternately. Moreover, two cities at the bottom are partly overlapped. Thus, there are additional paths in the upper right corner of the image and the paths are fuzzy at the bottom. In d, there are four center cities resulting in an extremely concave shape for the TSP loop; moreover, two of the cities are very close. Affected by these cities, the FCN misjudges the path at the bottom and hesitates on deciding paths related to the two close cities. In spite of this, the deviation between the produced solution and the optimal solution has been controlled to a low level. Samples like c and d are only a minority. The above experiments prove that the FCN is capable of ensuring good quality solutions for new TSP samples. Detailed analysis and discussions are further presented as follows.

**5.1 Image representation of input**

The key to this project is to convert TSP to fully connected graphs. In the graph, the locations of the cities can be directly converted according to the cities' coordinate set, while the paths among the cities are added deliberately, which greatly contributed to the solving abilities of the FCN. Actually, there are other ways to represent a TSP with an image. For example, a scatter image denoting each city as a dot in a map also fully describes the original information. But without the path information, it is difficult for the neural network to generate the paths. We also tested the performance of FCN with scatter images as the input. FCN is trained using the same training samples. After the same number of training iterations, we use the same test samples to check the network's solution quality. The results are listed in Table 2. The table shows that when paths are removed from the input, the network's ability to solve TSP worsens significantly. The portion of samples that can obtain the optimal solution is less than 50%. Even if the threshold is increased to 10%, only 90.83% results are within tolerance.

**Table 2 Metrics on test samples with a scatter image as input**

| $e_0$ | $e_1$ | $e_2$ | $e_5$ | $e_{10}$ | $R_{aver}$ |
|---|---|---|---|---|---|
| 48.33% | 58.13% | 66.67% | 80.00% | 90.83% | 102.85% |

A typical example is shown in Fig. 5 to compare the two image representation methods. The importance of the paths is mainly reflected in the fact that they constitute the superstructure of the TSP, which contains all the feasible solutions. When the paths are removed from the image, the relevant information of the optimal solution is also removed. It seems easier for the FCN to do "subtraction" than "addition". The superstructure in the fully connected graph directs the search space of the FCN, which just selects a best path loop among all feasible loops. With a scatter image as the input, the FCN needs to "create" the optimal path by itself based on the information of city positions, which greatly increases the complexity of the task. By using the scatter image, many solutions that do not meet TSP's constraints will be given by the FCN, indicating that expansion of the search scope will make it harder for the network to learn. In addition, due to the absence of "line" information in the input, the quality of the lines generated by the FCN significantly worsen, and some lines even bend.

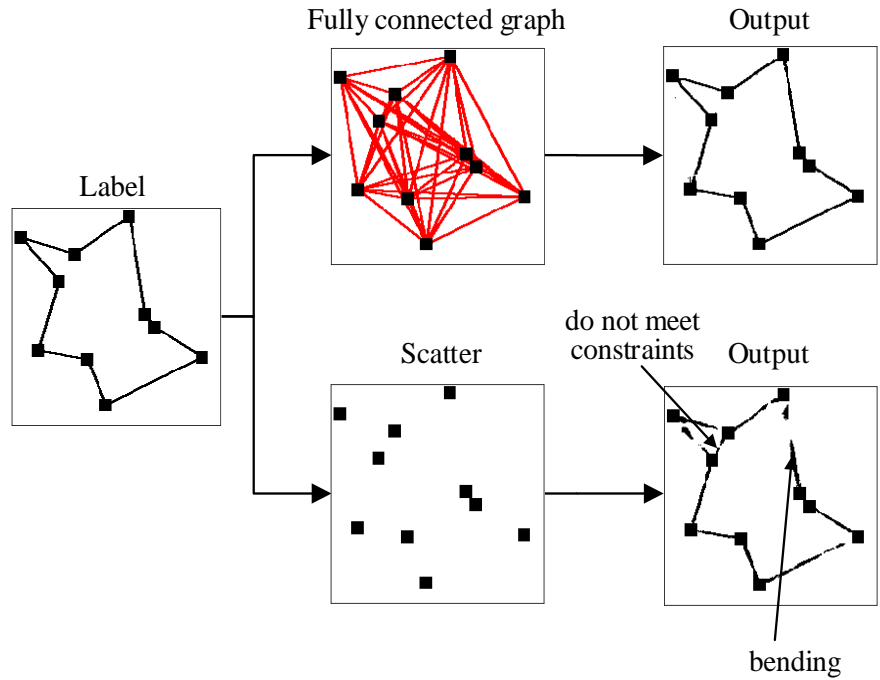

**Fig. 5. Solutions based on different image representation of input.**

**5.2 Post processing effect**

In section 4.3, a post processing algorithm is developed to automatically retrieve solutions of the FCN. As shown in Algorithm 1, there is a key parameter $m$, denoting how many cities are repeatedly selected as the departure city in post processing. To accurately measure the improvement and time overhead of the post processing algorithm, we compare the performance with different departure city parameters $m$.

The same test samples of 10 cities are used in this experiment. We first determine the solution metrics and the average value of the processing time of the post processing algorithm with a fixed departure city. Subsequently, we randomly selected 2, 3, ...,10 cities from 10 cities with repetition to be the departure city in turn. Their corresponding metrics and time consumption are calculated. In Fig. 6, the solution metrics are illustrated by histogram, and the average processing time is shown in the line chart. Fig. 6 indicates that the post processing algorithm can obviously enhance the solution metrics. As the number of departure cities increases, the effect of the improvement becomes prominent, but the rate of improvement slows down, indicating that when the number of departure cities reaches a certain value, samples that have not been processed in the correct order are negligible. Moreover, for an improved greedy algorithm that selects *m* cities as departure cities, it needs to connect paths *mn(n-1)/2* times. Thus, the time consumption increases linearly with the increase of *m*, which is basically consistent with the experimental results. As the linear time overhead is still on the scale of milliseconds, it is worth spending the time to improve the accuracy. Thus, we set the parameter as 10 in the application. The results shown in Table 1 were obtained with this setting.

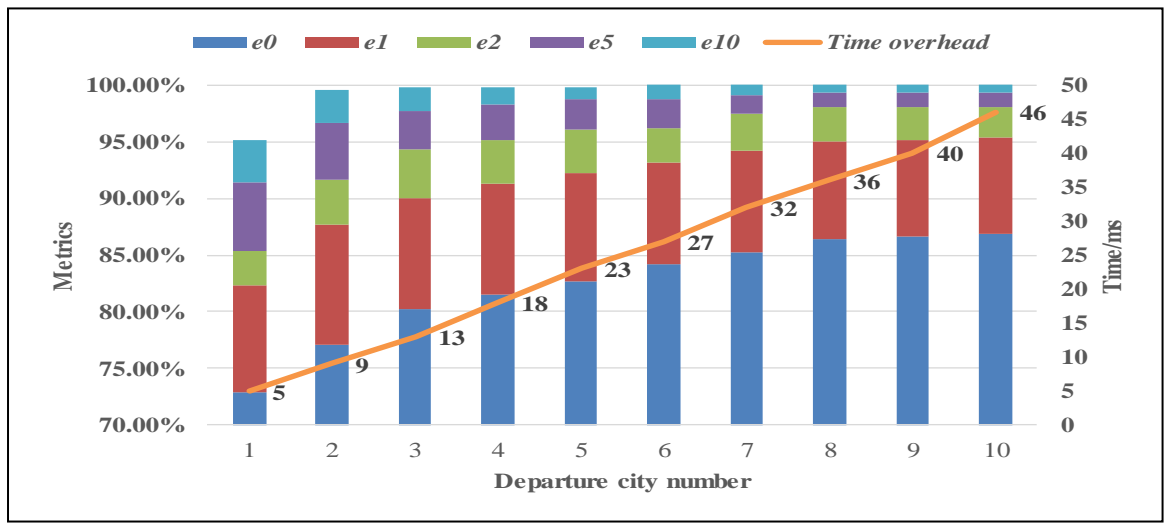

**Fig. 6. The solving metrics and the time consumption of post processing algorithms that adopt different departure city number.** The horizontal axis is the number of cities selected to be the departure city. The main vertical axis is the accuracy ratio and the secondary vertical axis is the time.

## 5.3 Generalization capability

In the previous experiments, TSP samples in 10 cities were used as training dataset. A FCN has shown great solving ability in predicting other new samples with changing city locations. Here, we will show that the FCN also performs well when applied to TSP instances with different numbers of cities. By varying the city number between 4 and 12, we generate 480 samples to validate the prediction model. The $e_0$ and $R_{aver}$ metrics on these samples are illustrated as in Fig. 7.

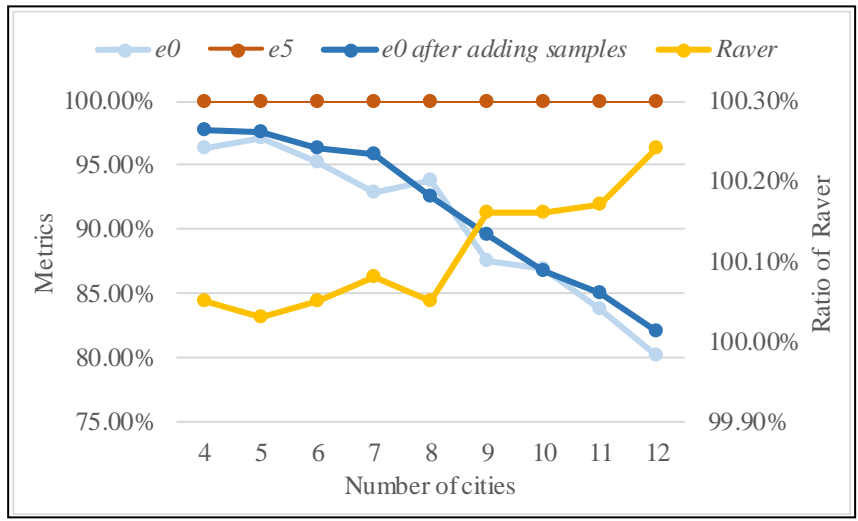

**Fig. 7. FCN's performance when solving TSP samples that have different city numbers.** The horizontal axis is the number of cities. The main vertical axis is the accuracy ratio and the secondary vertical axis is ratio of $R_{aver}$.

As can be seen from the figure, when the number of cities is less than 10, the network achieves higher $e_0$ metric than the metric for 10 cities, and the $R_{aver}$ metric is lower than the metric of 10 cities, which all indicate that the FCN performs better on these samples than in the 10 city case, though no samples with such a city number have been training before. The reason may be that as the number of cities is reduced, the image features existing on the picture and the feasible space also reduce. It reveals that the FCN in some way understands the task of solving a TSP.

When the number of cities exceeds 10, though the e0 metric drops slightly and the $R_{aver}$ metric increases slightly, the performance of the FCN is still stable. In the samples of 12 cities, above 80% of the samples can still reach the optimal solution, and $R_{aver}$ stays below 100.5%. This shows that the slight increase of city number does not bring too much hindrance to the performance of the FCN. Furthermore, the FCN's performance can be improved by adding a small number of samples of the corresponding city number for transfer learning [22]. As shown in Fig. 7, by adding only 1500 samples, the performance of FCN can be further enhanced. The accuracy on 12 cities can be improved to 82.08%. These results prove that the derived FCN has good generalization capability

### 5.4 Comparing with other algorithms

Table 3 The time consumption of several algorithms

| Number of cities | Exact algorithm | | | Approximate algorithm | | | | Proposed method | | |
|---|---|---|---|---|---|---|---|---|---|---|
| | Exhaustive Search | Dynamic Programming | Brand and Bound | Genetic population=300 cross rate=0.85 | | Ant Colony antNum=8 $\rho = 0.5$ | | FCN | Post processing | $e_0$ |
| | | | | Time | $e_0$ | Time | $e_0$ | | | |
| 4 | 0.06 | 0.51 | 0.73 | 166.59 | 100.0% | 5.53 | 99.17% | **6.60** | **2.65** | 96.25% |
| 5 | 0.21 | 0.71 | 2.63 | 170.11 | 100.0% | 8.01 | 98.75% | **6.54** | **5.76** | 97.08% |
| 6 | 0.87 | 3.40 | 10.81 | 173.73 | 100.0% | 21.98 | 97.71% | **6.55** | **10.13** | 95.21% |
| 7 | 6.13 | 19.88 | 33.45 | 178.38 | 100.0% | 42.68 | 94.79% | **6.89** | **16.14** | 92.92% |
| 8 | 57.40 | 143.36 | 103.42 | 181.62 | 100.0% | 71.12 | 93.54% | **6.72** | **23.87** | 93.75% |
| 9 | 581.60 | 1448.85 | 571.18 | 185.69 | 100.0% | 131.75 | 92.71% | **6.97** | **34.02** | 87.50% |
| 10 | 6563.20 | 10514.53 | 1347.00 | 316.90 | 99.79% | 185.08 | 89.79% | **6.73** | **46.49** | 86.88% |
| 11 | 223405.60 | 101818.27 | 6933.23 | 325.07 | 98.12% | 417.43 | 88.75% | **6.61** | **60.72** | 83.75% |
| 12 | 2420217.00 | 632438.10 | 14603.91 | 730.39 | 95.42% | 439.41 | 86.86% | **6.88** | **78.30** | 80.21% |

Note: The unit of time is ms

Other typical deterministic algorithms and heuristic algorithms are also used to solve the TSP samples as comparisons. Table 3 lists all the results. We find that the deterministic algorithms solve fast for small city numbers; but as city number increases, the time consumption increases dramatically. In contrast, the heuristic algorithms take more time for small scale problem; but with the increase in city number, the time cost increases steadily.

When using the FCN, no matter how many cities, the time consumption is around 6.5-7ms, because the FCN only needs to forward propagate to derive the answer after training. Though the post processing costs more time, the overall time costs are in milliseconds, which can guarantee real-time application. Moreover, when dealing with a new TSP application, the other methods must run the optimization repeatedly. The proposed method, however, does not need re-application of an iterative method.

## 6 CONCLUSION

In this paper, a FCN is introduced to solve the TSP, a typical NP-hard optimization problem. Instead of adopting iterative searching, the FCN learns the mapping from all feasible solutions to optimal solution. The key to this project is adopting an image representation, which enables the FCN to extract useful information from the TSP parameters. After training, the FCN can be directly applied to predict solutions to new TSP samples without re-iteration. Experiments show that high solving accuracy can be obtained through the FCN. In addition, FCN also has great generalization capability. We finally show that the time consumption of the proposed method on new problems is much less than other optimization algorithms.